\documentclass[conference]{IEEEtran}
\ifCLASSINFOpdf
  \usepackage[pdftex]{graphicx}
  \DeclareGraphicsExtensions{.pdf,.jpeg,.png}
\else
\fi

\begin{document}

\title{Spiking Neural Streaming Binary Arithmetic}

\author{\IEEEauthorblockN{James B. Aimone, Aaron J. Hill, William M. Severa, $\&$ Craig M. Vineyard}
\IEEEauthorblockA{Sandia National Laboratories  \\
Albuquerque, New Mexico \\
Email: jbaimon@sandia.gov}
}

\maketitle

\begin{abstract}
Boolean functions and binary arithmetic operations are central to standard computing paradigms. Accordingly, many advances in computing have focused upon how to make these operations more efficient as well as exploring what they can compute. To best leverage the advantages of novel computing paradigms it is important to consider what unique computing approaches they offer. However, for any special-purpose co-processor, Boolean functions and binary arithmetic operations are useful for, among other things, avoiding unnecessary I/O on-and-off the co-processor by pre- and post-processing data on-device.  This is especially true for spiking neuromorphic architectures where these basic operations are not fundamental low-level operations.  Instead, these functions require specific implementation.  Here we discuss the implications of an advantageous streaming binary encoding method as well as a handful of circuits designed to exactly compute elementary Boolean and binary operations. 
\end{abstract}

\IEEEpeerreviewmaketitle

\section{Introduction $\&$ Background}
Fundamental to many paradigms of computing are Boolean functions and arithmetic operations. These core concepts can then be composed to build arbitrarily complex computations and set a foundation for comparing implementations and understanding computability. In pursuing an understanding of what computations neural circuits can perform, prior work has explored universal function approximation as well as Turing completeness \cite{cybenko1989mathematics,schuman2021neuromorphic}. Accordingly, with that foundation in hand it is straightforward that spiking neurons can be used to compute arithmetic functions. However, here we not only provide several fundamental arithmetic computations as spiking neural streaming circuits, but use them as a means of understanding and enabling neuromorphic computing (NMC). Accordingly, here we present a set of streaming neural binary circuits implemented in Fugu, a neural algorithm composition framework, showing how more complex functions can be built upon operations such as addition and subtraction leading to multiplication.  

Classic computational paradigms are incredibly efficient at performing these base building blocks of numerical computation, having been optimized for decades to minimize the computational kernel and maximize scalability \cite{hennessy2017computer,hennessy2011computer}. Exacting these computations in neurons both shows potential for how future, device breakthroughs in the development of neuromorphic hardware can enable classic numerical computations. 
And this work has been inspired partly by previous approaches for implementing logic and arithmetic in spiking networks, such as~\cite{lagorce2015stick,reljan2017solving,sahni2019implementation,maass1997networks}. 
But furthermore, this exploration also examines how the computational flexibility in spiking neural networks can be leveraged to enable compositionality for more complex computations. Alternatively, if the highly optimized canonical approaches were used to compute fundamental arithmetic operations which integrate neural sub-functions, there is a cost to convert in and out of neural circuity analogous to paying for analog to digital conversions.   

\section{Fugu}
As a means of showing compositionality and scalability of spiking algorithms, we use the Fugu framework to represent the neural circuits presented here \cite{aimone2019composing}. While implementation details vary based on NMC hardware, Fugu is a high-level framework specifically designed for developing spiking circuits in terms of computation graphs. Accordingly, with a base leaky-integrate-and fire (LIF) neuron model at its core, neural circuits are built as `bricks'. These foundational computations are then combined and composed as `scaffolds' to construct larger computations. This allows us to describe the streaming binary arithmetic circuits in terms of neural features common to most NMC architectures rather than platform specific designs. 

In addition to architectural abstraction, the compositionality concept of Fugu not only facilitates a hierarchical approach to functionality development but also enables adding pre and post processing operations to overarching neural circuits. Such properties position Fugu to help explore under what parameterization or scale a neural approach may offer an advantage. For example, prior work has analyzed neural algorithms for computational kernels like sorting, optimization, and graph analytics identifying different regions in which a neural advantage exists accounting for neural circuit setup, timing, or other factors \cite{verzi2018computing, verzi2017optimization, parekh2018constant, hamilton2018neural}. 

\section{Spiking Binary Arithmetic}
An open research question in neuroscience and neuromorphic computing is how to encode information \cite{schuman2019non}. The transmission of spikes can conveys information in their timing, enabling complex spatial temporal representations. However, here we do not exploit any novel spike encoding representations but rather use a binary representation of numbers which starts streaming the least significant bit first to the most significant bit last. Neurons can be used to represent many different coding schemes, but the main advantages of this ``little-endian'' temporal binary representation are that:
\begin{itemize}
   \item One neuron is required per variable represented, with $k$ timesteps required for a $k$-bit number.
   \item Overflow can be handled by simply increasing the length of the vector (adding one additional timestep).
   \item We can construct streaming addition, subtraction, and multiplication based on standard binary arithmetic operations.
\end{itemize}

The circuits described here are intended to be compatible with any large-scale NMC platform, which means that we mostly avoid using decays as they can be additive or multiplicative depending on the architecture. We note exception to this in our discussions, such as in the inequality check in section \ref{sec:inequality}.

\subsection{Streaming Adder}
\begin{figure}[!t]
\centering
\includegraphics[width=3in]{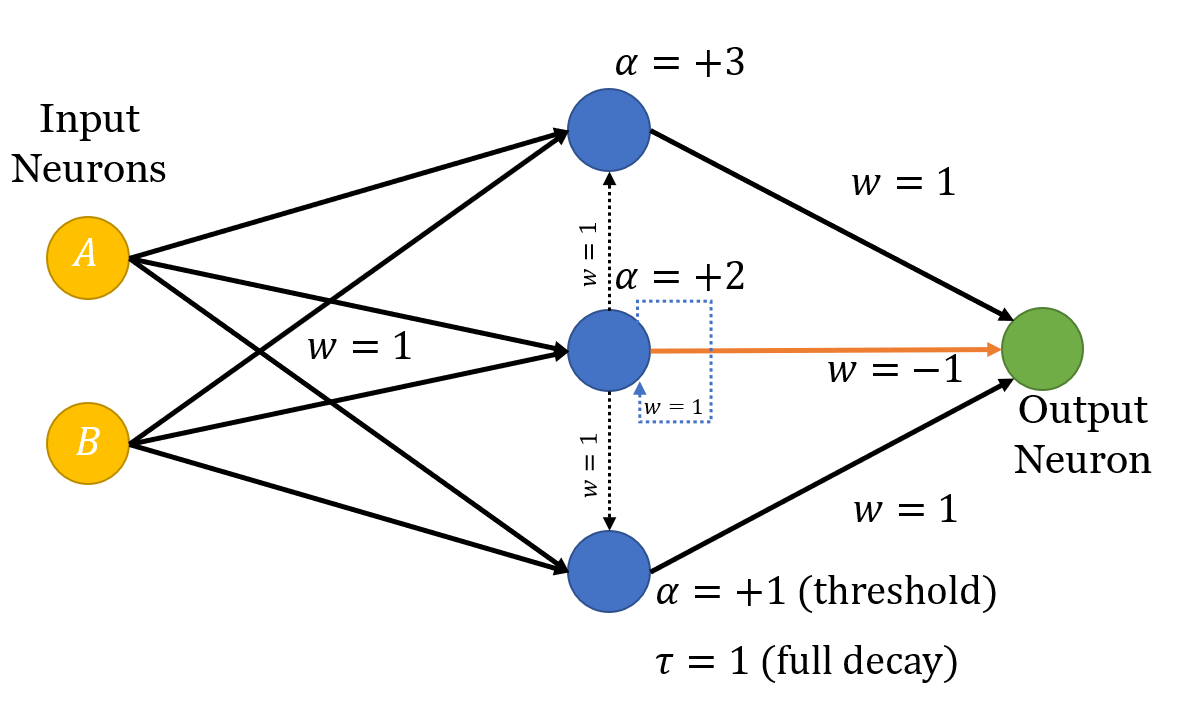}
\caption{Schematic of a binary adder circuit. This circuit is implemented as a `brick' in Fugu, allowing it to be re-used in more complex circuits}
\label{binary_adder}
\end{figure}

One fundamental operation that we can easily implement is an adder, which we have implemented in Fugu, the IBM's TrueNorth Architecture \cite{merolla2014million}, and the Intel Loihi \cite{davies2018loihi} platform.  An adder is an instrumental function in digital electronics. Besides its use in arithmetic logic units directly, the adder operation is involved in many control tasks of a processor as well. At the bit level, this operation simply produces the outcome of combining bits for all possible combinations. This simple concept becomes slightly more complex when needing to handle overflow or carry. But foundationally, the concept can readily be implemented as a series of Boolean gates which account for the number of bits received. The connectivity, thresholds, and decay dynamics of the neural circuit are constructed such that the neurons can implement logic functions. For example, with a threshold of one, a neuron spikes upon receiving any input (analogous to a logical OR). Decay dynamics impose structure upon when inputs must arrive. And the internal state of the neurons allows the circuit to store a carry value. Altogether, these operations enable the neural circuit to route the flow of spikes, much like a canonical binary full adder does using standard logic gates. The adder that we constructed takes two inputs (denoted $A$ and $B$ in Figure \ref{binary_adder}) and uses three hidden neurons with different thresholds.

\subsection{Inversion}
To support a subtraction operation we need to invert a binary stream which is equivalent to a NOT gate. Our implementation of a NOT gate on TrueNorth requires one neuron and is depicted in Figure \ref{fig:support_circuits}a. Our binary number representation in the spike domain is that a spike in time represents a 1 and no spike in time represents a 0. The NOT gate neuron inhibits any incoming spikes, but if no spike is received the neuron will spike. The implementation details for this will vary based on the NMC hardware being used. For TrueNorth, we configure a neuron to have a positive leak value of $1$ and a threshold of $1$. The input synapse weight to the neuron is set at $-1$. In this configuration, if no spike is received, the positive leak value will cause the neuron to cross threshold and spike at every time step, resetting its neural potential to $0$. However, if a spike is received the synapse weight of $-1$ will be added to the leak value of $+1$ and the neuron potential will be unchanged\footnote{~This effect is a result of the specific nature of the TrueNorth neural dynamic in which the leak is applied to the stored potential before the threshold check. Other NMC platforms will require a different neuron configuration.} causing no spike to occur. When this type of neuron dynamic receives a stream of spikes that represents a binary number, the output of the neuron will be the bitwise negation of the input stream. On platforms with multiplicative leak, such as Loihi, a comparable circuit can be generated that requires an extra neuron.  

\begin{figure}[!t]
\centering
\includegraphics[width=2.75in]{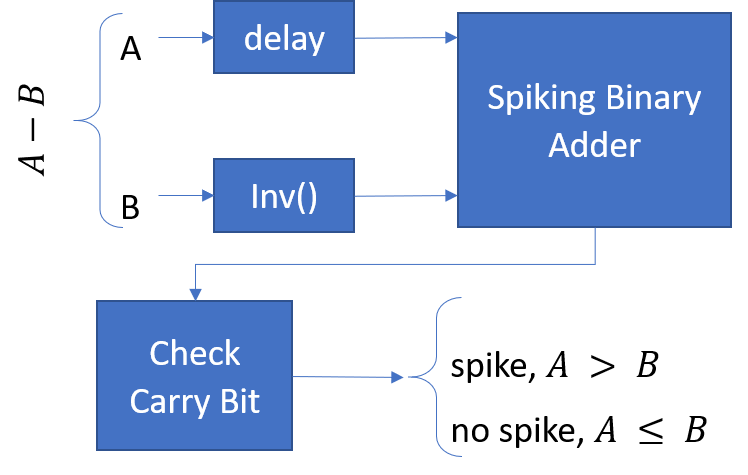}
\caption{Inequality Check Circuit leveraging the binary spike adder with an inversion neuron and carry bit check circuit. Each box consists of a separate brick within Fugu.}
\label{fig:inequality}
\end{figure}

\subsection{Inequality}
\label{sec:inequality}
The streaming adder can serve as the basis for building more complex arithmetic with the addition of small support circuits that implement additional base-logic functions. One such support circuit described above is the inverting neuron, or NOT gate. With this inverting neuron we can perform an inequality check (Figure \ref{fig:inequality}) that solves the simple logical question of is $A$ greater than $B$, for two unsigned integers $A$ and $B$. We do this by leveraging the binary ones compliment subtraction method. In binary ones compliment representation a negative integer is obtained by inverting each bit of its positive value representation. To use this to solve the inequality check of determining if $A > B$, we perform the operation of $A - B$ in ones compliment representation and use the ``end around carry'' bit as a signal to determine the result. It can be shown that for unsigned integers $A$ and $B$, the ones compliment subtraction method, $A - B = A + inv(B)$, will produce a carry bit value of $1$ if $A > B$ and $0$ if $A \leq B$. We restrict our inputs to unsigned integers to avoid the complexities of overflow, since the subtraction of two positive $N$-bit integers cannot produce a result outside the range of the $N$-bit representation.

\begin{figure}[!t]
\centering
\includegraphics[width=2.5in]{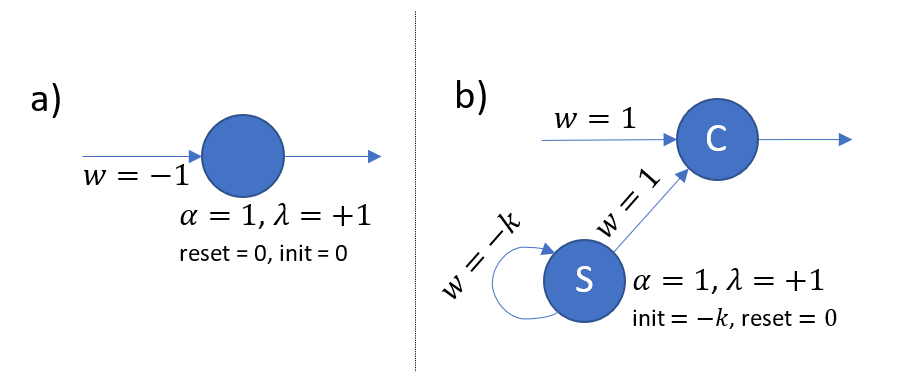}
\caption{Supporting neural logic circuits for platforms with additive leak such as TrueNorth. (a) inverting neuron, (b) check carry bit circuit. }
\label{fig:support_circuits}
\end{figure}

For our inequality check circuit (Figure \ref{fig:inequality}), we are computing the expression $A + inv(B)$ and checking the final carry bit to determine if the expression $A > B$ is true or false. After leveraging the adder and NOT gate to perform the computation of $A + inv(B)$ we produce the result by inspecting the carry bit. This inspection requires an additional support circuit requiring two neurons (Figure \ref{fig:support_circuits}b). The main neuron of this circuit, the check neuron, receives the result of the adder and will spike if the final carry bit of the adder result is $1$ and will not spike for any other input. To do this we create an additional input to the check neuron from a self-spiking neuron that provides a spike to the check neuron at the exact time the final carry bit of the adder result is entering the check neuron. This precise timing is maintained by a spike back to itself with a weight of $-k$, which will reset its potential back to $-k$.

Depending on the manner the neuromorphic hardware implements leak, there is an alternative approach to inequality that uses only one neuron if there is multiplicative leak, such as on Loihi. The check neuron receives a sub-threshold positive input from $A$ and a negative input from $B$, and uses a $0.50$ multiplicative decay at each timestep. After both inputs are complete, the check neuron can be interrogated (by an input at exactly threshold), to determine if the cumulative voltage is greater than zero, thus $A>B$, or less than zero, thus $A<B$ (The circuit can be modified to check for equality as well). This approach is considerably more efficient, but it is not universal on NMC platforms, and further has the drawback that unlike the arithmetic circuit above, its intermediate computation is not as useful for other operations, and it relies on sufficient numerical precision within the neuron's voltage representation.

\begin{figure}[!t]
\centering
\includegraphics[width=2.5in]{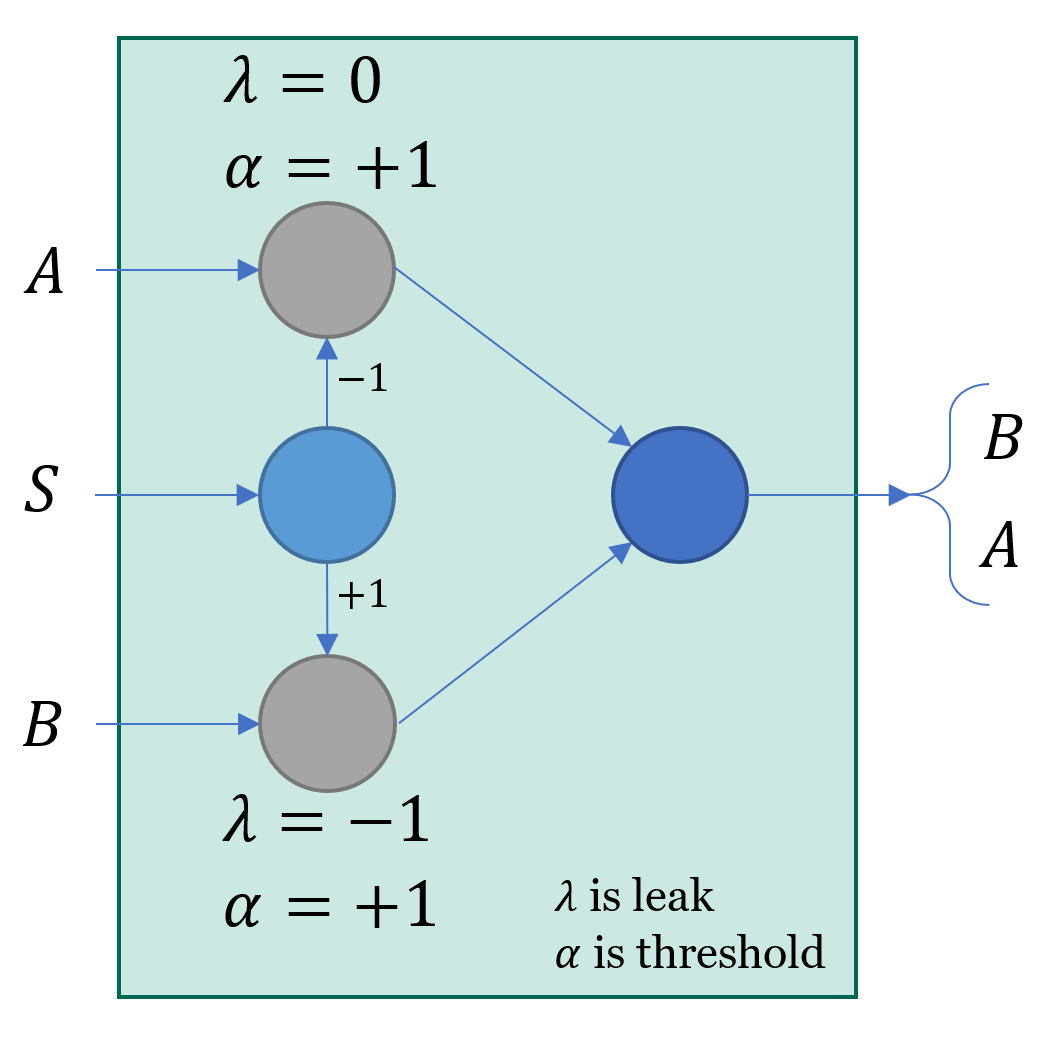}
\caption{Spiking 2-to-1 Multiplexer: The $A$ input neuron is configured to pass spikes forward if there is no input from the select neuron, while the $B$ input neuron is configured to inhibit any incoming spikes if there is no input from the select neuron. This functionality is reversed if the select neuron begins spiking.}
\label{fig:mux}
\end{figure}

\begin{figure*}[!t]
\centering
\includegraphics[width=\textwidth]{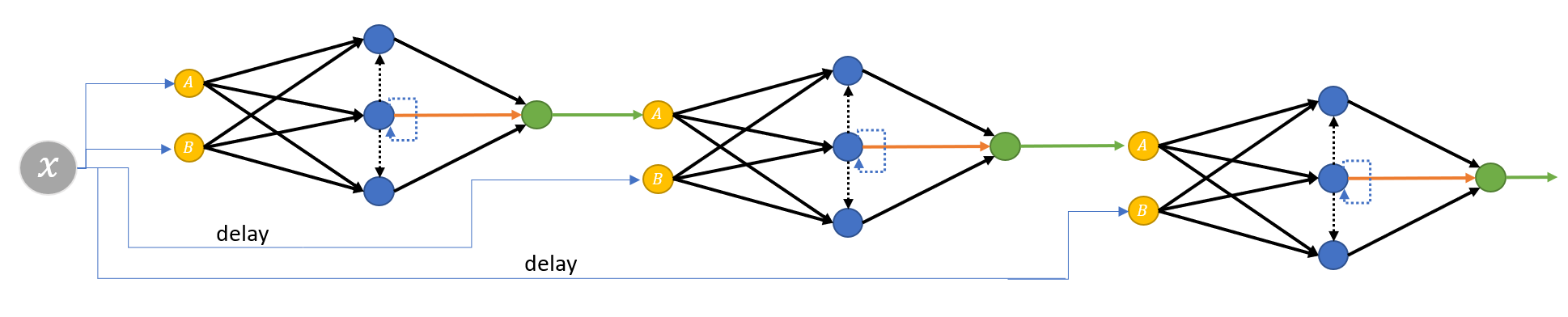}
\caption{The cascading of multiple adders combined with delays can compose a fixed multiplication circuit. The multiplier can be natively implemented in Fugu by combining pre-defined bricks for the adder and temporal delay circuits.}
\label{fig:fugu_mult}
\end{figure*}

\subsection{Min/Max}
A min or max operation can be performed by utilizing the inequality check circuit. The result of the inequality check circuit provides a signal that identifies which input is larger or smaller. This is the same computation that a min or max function would need to perform. What is missing is the ability to select the full input of $A$ or $B$ and pass it down stream for later use. For this we develop a streaming spike based 2-to-1 multiplexer (mux). Here the inputs of $A$ and $B$ are split first, with one copy entering the inequality circuit and the other copy being delayed and fed into the 2-to-1 mux. The result of the inequality check circuit is fed into the select input of the mux. That is, if the inequality check circuit spikes it is indicating that $A > B$, signaling the mux to pass the stream of spikes on the $A$ input to perform a max operation or pass the stream of spikes on the $B$ input to perform a min operation. The 2-to-1 mux is a carefully crafted neural circuit whose implementation details will depend on the underlying NMC hardware being used. We have implemented this circuit on TrueNorth but we will only conceptually describe its function here.

The spike based 2-to-1 mux requires four neurons, as depicted in Figure \ref{fig:mux}. Two neurons for each input $A$ and $B$, one select neuron, and one output neuron. When the select neuron is providing no input, one input neuron will be inhibited and the other will pass its input forward. The select neuron is crafted to receive the input of the inequality circuit and will either not spike or will sequentially spike for each of k timesteps, where k is the bit length, and then stop until another input is received. Because the select input is connected to each input neuron with either a positive or negative weight, when the select neuron is excited it reverses the function of the input neurons. In Figure \ref{fig:mux}, an excited select neuron would cause the A input neuron to inhibit, and the B input neuron to mirror the input B through to the output neuron. The output neuron passes any spike it receives.

\subsection{Subtraction}
The functional blocks described thus far can be used to implement full binary subtraction. This is done by understanding that when performing binary subtraction, we use the addition form, that is $A - B = A + (-B)$. This requires a signed binary representation. The standard signed binary representation is twos compliment. It is possible to implement a full subtraction in ones compliment form, but it will not be detailed here\footnote{~The basis of a ones compliment subtraction was provided when describing the inequality check circuit.}. Given a binary number $B$, the negation of that value in twos compliment representation is performed by inverting all the bits and adding one. That is $-B = inv(B) + 1 = 2comp(B)$. Therefore, $A - B = A + (-B) = A + (inv(B) + 1)$, in twos compliment representation. By using our inverting neuron and adder circuit we can perform the $inv(B) + 1$ computation and feed this result into another adder that receives $A$ at the appropriate time, thus performing $A  - B$ in twos complement representation.

\subsection{Scalar Multiplication}
Lastly, we describe a streaming scalar multiplier to perform the function $y=a*x$, where $x$ is a variable in our temporal binary format and $a$ is a pre-programmed scalar of some precision. The multiplication of two numbers in binaries is equivalent to the process of long multiplication in decimal. However, since in binary each element equals either zero or one, multiplication becomes a series of additions and multiplications times powers of $2$. What is convenient about the streaming little endian coding scheme used here coupled with neuromorphic hardware is that we can multiply by powers of $2$ by implementing a simple bitshift of the variable $x$, which is simply adding a timestep delay, to multiply the variable by $2$. So scalar multiplication of $a$ times $x$ amounts to cumulatively adding bit-shifted versions of $x$ for each element of $a$ that equals $1$. Viewed as a combination of additions and temporal-shifts, scalar multiplications becomes a natural fit to Fugu. We now take two bricks (the above adder brick, and a simple temporal shift brick) and we can instantiate a straightforward scalar multiplication (Figure \ref{fig:fugu_mult}). Though it adds complexity, using an inhibitory signal to gate certain adders allows this scalar multiplication to be extended from a fixed $a$ times $x$ to two variables $x$ times $y$.

\section{Conclusion}
While much research in neural computation has been for data science tasks, there is also value in exploring numerical computing and compound functions. Surely this approach is roundabout and will not replace all forms of computing.  However we see value in specific uses applicable both for the existing computing hardware of today but also emerging paradigms. This perspective of function composition and streaming binary representation offers the ability to effectively increase the scope and applicability of tasks appropriate for neuromorphic processors.  Furthermore, these seemingly straightforward approaches are due to our choice in spiking representation and represent the benefits advantageous spike encoding can confer.

Given that today's large-scale neuromorphic hardware leverages digital implementations of neurons who typically have arithmetic units inside, why would one use neurons to perform these basic computations?  One immediate reason is the cost of I/O on large scale neuromorphic platforms. Because data movement on and off of a chip is orders of magnitude more expensive than on chip communication (and in many cases not possible given the designs), it is critical to develop basic tools to analze data in situ. For instance, in machine learning algorithms, identifying a neuron that is active the most over a data set may have value, so using simple arithmetic to perform that cumulative count and comparison on chip is advantageous, despite costing a few neurons.  

Longer term there is a value to these arithmetic circuits when considering the trajectory of more powerful devices.  As the microelectronics community moves away from a primary focus on transistor like behavior, one possible outcome are devices that naturally perform more threshold-gate like computation (e.g., a tunable threshold, multiple inputs).  In that case, as device level logic increases in complexity, even modestly, these circuits described here may have tremendous value.

\newpage
\section*{Acknowledgment}

Sandia National Laboratories is a multimission laboratory managed and operated by National Technology $\&$ Engineering Solutions of Sandia, LLC, a wholly owned subsidiary of Honeywell International Inc., for the U.S. Department of Energy’s National Nuclear Security Administration under contract DE-NA0003525.  

This paper describes objective technical results and analysis. Any subjective views or opinions that might be expressed in the paper do not necessarily represent the views of the U.S. Department of Energy or the United States Government.

SAND2021-13472 C


\end{document}